\documentclass[10pt,twocolumn,letterpaper]{article}

\usepackage[pagenumbers]{wacv} 

\usepackage{graphicx}
\usepackage{amsmath}
\usepackage{amssymb}
\usepackage{booktabs}
\usepackage{graphicx}

\usepackage{float}

\usepackage{booktabs}
\usepackage{adjustbox}
\usepackage{textcomp}
\usepackage{multirow}
\usepackage{xcolor}
\usepackage{colortbl}
\usepackage{todonotes}
\usepackage{enumitem}
\usepackage{footmisc}
\usepackage{amsmath}
\usepackage{comment}
\usepackage{tabularx}
\usepackage{listings}
\usepackage{graphicx} 
\usepackage{subcaption} 
\usepackage{array}
\usepackage{booktabs}
\usepackage{siunitx}

\definecolor{ao(english)}{rgb}{0.0, 0.5, 0.0}
\definecolor{cornellred}{rgb}{0.7, 0.11, 0.11}
\definecolor{pf7}{RGB}{166, 118, 29}

\newcommand{\furl}[1]{\footnote{\url{#1}}}

\usepackage[pagebackref,breaklinks,colorlinks]{hyperref}
\usepackage{enumitem}

\usepackage[capitalize]{cleveref}
\crefname{section}{Sec.}{Secs.}
\Crefname{section}{Section}{Sections}
\Crefname{table}{Table}{Tables}
\crefname{table}{Tab.}{Tabs.}

\makeatletter
\AtBeginDocument{
	\let\@oldmaketitle\@maketitle
	\renewcommand{\@maketitle}{\@oldmaketitle
		\vspace{-15pt}
		\centering
		\includegraphics[width=\linewidth]{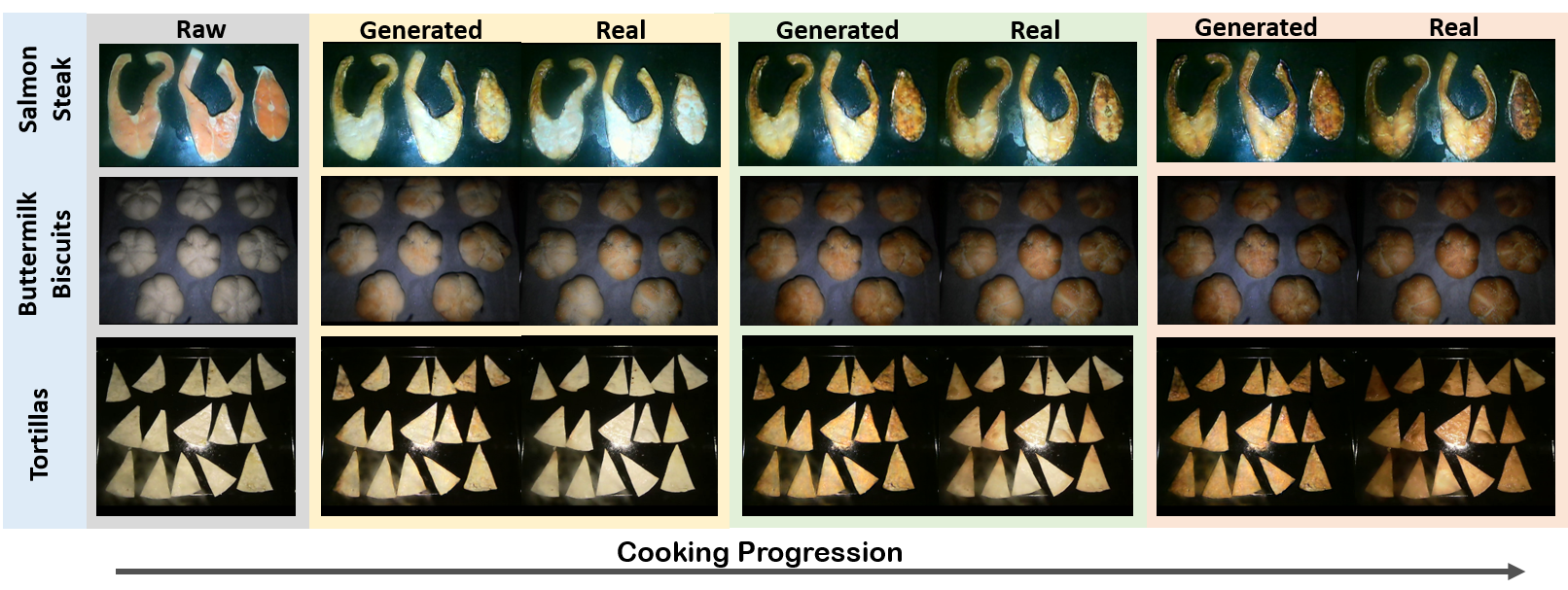}
		\vspace{-10pt}
		\captionof{figure}{
			Given \textbf{raw} food images and text prompts \textbf{(recipe name, cooking state)}, our model generates three visually distinguishable, realistic images for each cooking state showing cooking progression. Best viewed in color.
		}
		\label{fig:teaser}
		\vspace{17pt}
	}
}
\makeatother

\begin{document}

\title{Real-Time Cooked Food Image Synthesis and Visual Cooking
Progress Monitoring on Edge Devices
}

\author{
	Jigyasa Gupta,
	Soumya Goyal,
	Anil Kumar,
	Ishan Jindal\\
	Samsung R\&D Institute India, Delhi\\
	{\tt\small \{jigyasa.g,soumya.goyal,anil.k06,ishan.jindal\}@samsung.com}
}

\maketitle

\begin{abstract}
Synthesizing realistic cooked food images from raw inputs on edge devices is a challenging generative task, requiring models to capture complex changes in texture, color and structure during cooking. Existing image-to-image generation methods often produce unrealistic results or are too resource-intensive for edge deployment. We introduce the first oven-based cooking-progression dataset with chef-annotated doneness levels and propose an edge-efficient recipe and cooking state guided generator that synthesizes realistic food images conditioned on raw food image. This formulation enables user-preferred visual targets rather than fixed presets. To ensure temporal consistency and culinary plausibility, we introduce a domain-specific \textit{Culinary Image Similarity (CIS)} metric, which serves both as a training loss and a progress-monitoring signal.  Our model outperforms existing baselines with significant reductions in FID scores (30\% improvement on our dataset; 60\% on public datasets).
\end{abstract}

\section{Introduction}
Cooking induces complex, non-linear visual and physical transformations, such as browning, caramelization, bubbling, texture change, that serve as primary indicators of doneness in everyday culinary settings. These appearance-based cues are often more reliable than time or internal temperature, particularly for foods where surface characteristics define quality. Nevertheless, most consumer ovens still rely on fixed timers or static thermal presets. 
Thermal probes, while effective for some meats, are unsuitable for common oven-prepared items such as bread, pastries, pizza crusts, and roasted vegetables. This gap between visual intent (“stop when golden brown”) and appliance execution (``90s at 700W'') motivates vision-based cooking feedback.

The increasing integration of cameras and low-power NPUs into consumer appliances \cite{june2020cooking,ge2020cooking,samsung2023cooking,lg2024cooking} enables real-time visual assessment on-device. Yet diffusion models \cite{rombach2022high,wang2025cookingdiffusion} remain computationally prohibitive for embedded deployment ($>$1GB)\footnote{Stable Diffusion V1.5 $\approx$983M parameters; Cooking Diffusion builds on this base}, and lightweight GAN-style models  \cite{isola2017image} struggle to capture fine-grained, progressive cooking transitions. Moreover, no existing benchmark models the temporal visual evolution of food inside an oven; prior datasets emphasize stovetop actions (YouCook2 \cite{ZhXuCoAAAI18} and COM Kitchens \cite{maeda2024comkitchensuneditedoverheadview})  or static food understanding (ISIA Food-500 \cite{min2020isiafood500datasetlargescale}) rather than appearance progression under heat.

We introduce the first benchmark and task definition for oven-based cooking progression, collected in controlled settings with chef-annotated cooking state. We define two coupled tasks: \textbf{(1) Real-Time Cooked Food Image Synthesis} - generating the expected cooked appearance conditioned on raw input, recipe name (e.g. ``Salmon Steak''), and doneness preference(e.g., ``Medium-rare''); and \textbf{(2) Visual Cooking Progress Monitoring}, stopping cooking when live observations align with the synthesized target. This formulation parallels human judgment, where desired outcomes are instance-dependent and visually grounded.

Direct doneness classification or remaining-time regression appears simpler, but our experiments show such models generalize poorly due to variation in food thickness, moisture, seasoning, and oven dynamics (results in Appendix \ref{appendix:doneness}). Further, scalar predictions offer limited interpretability or user preference control, whereas generation enables explicit doneness specification and visual transparency.

To this end, we propose an edge-efficient U-Net based generator guided by sinusoidal embeddings of recipe and cooking state. Sinusoidal embeddings, inspired by Transformer positional encodings, allow structured guidance across cooking states. Training leverages a novel \textit{Culinary Image Similarity (CIS)} loss, which, unlike SSIM or LPIPS, is trained on real cooking sequences. CIS enforces temporal consistency during generation and also supports progress monitoring by signaling when to stop cooking.  

Our contributions are threefold:
\begin{itemize}[noitemsep,topsep=0pt]
    \item We introduce a novel task and benchmark for cooked food image synthesis and real-time progress monitoring on edge devices. 
    \item We propose a compact text guided conditional image generator with a domain-specific similarity loss that captures culinary transformations
    \item We benchmark on a curated dataset of 1,708 cooking sessions across 30 recipes, achieving significant improvements over existing baselines (FID: 52 vs. 75, LPIPS: 0.2145 vs. 0.2523 on 715 image pairs) while running in real time (1.2s/image generation, 0.3s for image similarity on a 5 TOPS NPU). Our method also reduces FID by $>$60\% on public datasets (edge2shoes, edge2handbags).  
    \item 
\end{itemize}

The rest of this paper reviews related work in Section~\ref{sec:related}, presents our formulation in Section~\ref{sec:problem} and detailed experiments and results in Section~\ref{sec:results}.

\section{Related Work}
\label{sec:related}

Our work lies at the intersection of efficient image-to-image translation and domain-specific perceptual similarity, with the goal of synthesizing realistic cooked food images on resource-constrained edge devices.  

\textbf{Efficient image-to-image translation.}  
Pix2Pix \cite{isola2017image} introduced paired image-to-image translation using a conditional GAN with a U-Net generator and PatchGAN discriminator. While effective across tasks such as edge-to-photo, its one-to-one mapping is ill-suited to the one-to-many nature of cooking, where a single raw input can evolve into multiple plausible cooked states. SPADE \cite{8953676} extended control through spatially adaptive normalization, producing high-quality outputs but relying on 2D spatial layouts. This is different from our problem, where conditioning is global (recipe, cooking state) rather than spatial.  

Diffusion-based methods \cite{saharia2022paletteimagetoimagediffusionmodels,meng2022sdeditguidedimagesynthesis,kawar2023imagictextbasedrealimage,brooks2023instructpix2pixlearningfollowimage} have achieved state-of-the-art results in image editing and translation. Palette \cite{saharia2022paletteimagetoimagediffusionmodels} and SDEdit \cite{meng2022sdeditguidedimagesynthesis} adapt iterative denoising for image-to-image tasks, while text-guided methods like Imagic \cite{kawar2023imagictextbasedrealimage} and InstructPix2Pix \cite{brooks2023instructpix2pixlearningfollowimage} showcase fine-grained controllability. However, their iterative sampling is computationally expensive, making them impractical for real-time edge deployment, and often introduces stochastic artifacts.  

Pix2Pix-Turbo \cite{parmar2024onestepimagetranslationtexttoimage} proposed a one-step translation framework built on pre-trained diffusion priors, yielding fast and high-quality results. Yet, it inherits the heavy parameter count of its foundation model and suffers from occasional artifacts. Our work is motivated by the efficiency of Pix2Pix-Turbo but introduces a lightweight generator (8M parameters) explicitly designed for culinary transformations, enabling real-time edge inference.  

\begin{figure*}[!t]
	\centering
	\includegraphics[width=0.9\linewidth]{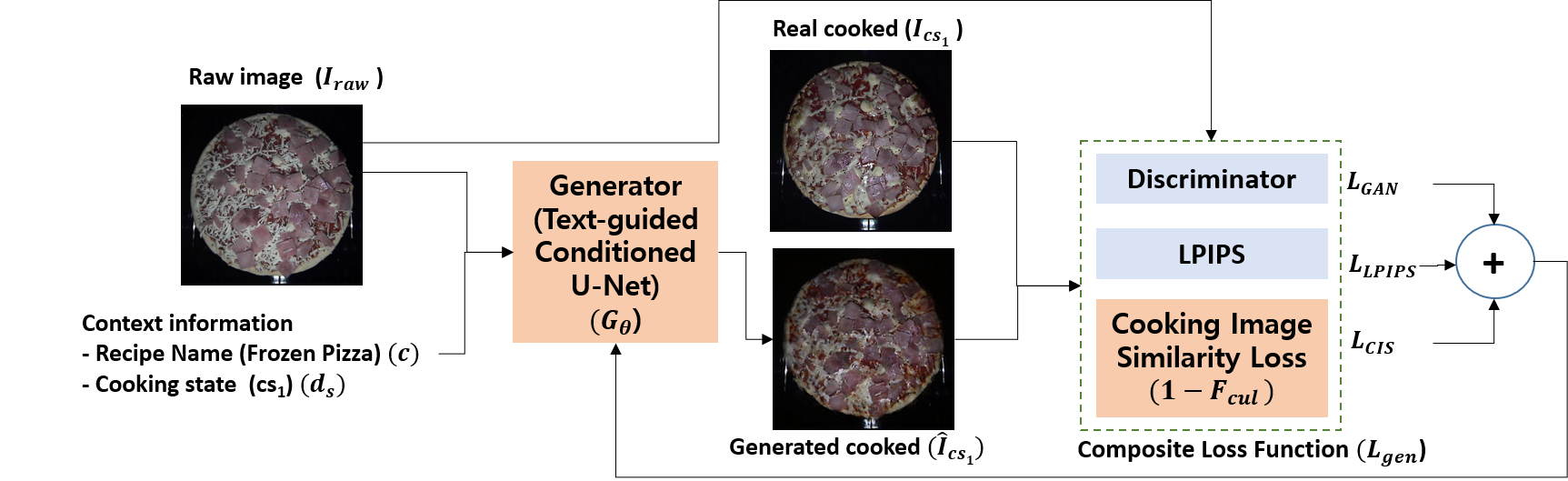}
	\caption[]{Overall generator architecture. The generator takes in input raw image and context information like recipe name and cooking state and generates a cooked image as output. The discriminator, employing a patch-based approach, evaluates both (raw,real-cooked) and (raw, generated-cooked) image pairs, providing the generator with adversarial loss. Additionally, the generator incorporates perceptual losses by comparing real-cooked and generated-cooked images using LPIPS and Culinary Image Similarity Loss. These three losses - adversarial, LPIPS, and CIS  are combined to optimize the generator during training}
	\label{fig:generatorarchitecture}
\end{figure*}

In summary, prior approaches range from classic GANs to powerful diffusion models, but none satisfy three requirements simultaneously: (1) handling  non-spatial text inputs such as recipe and cooking state, (2) lightweight design for edge deployment, and (3) high-fidelity, temporally consistent generation. Our edge-efficient text guided conditional generator is designed to meet these needs.  

\textbf{Culinary-aware perceptual similarity.}  
Image similarity metrics are critical for both training and progress monitoring. Traditional measures such as PSNR and SSIM \cite{wang2004image} poorly correlate with perceptual quality. Learned metrics such as LPIPS \cite{zhang2018unreasonable}, R-LPIPS \cite{ghazanfari2023rlpipsadversariallyrobustperceptual}, and LipSim \cite{ghazanfari2024lipsimprovablyrobustperceptual} leverage deep features for better alignment with human judgments, while SAMScore \cite{li2025samscorecontentstructuralsimilarity} exploits embeddings from large pre-trained models. Although effective for photorealism, these general-purpose metrics lack culinary awareness.  

We therefore propose the Culinary Image Similarity (CIS) metric, inspired by Siamese networks and contrastive learning \cite{mo2022siameseprototypicalcontrastivelearning,sermanet2018timecontrastivenetworksselfsupervisedlearning,chen2020simpleframeworkcontrastivelearning}. Trained on annotated cooking sequences, CIS learns an embedding space where progression unfolds as a smooth trajectory. Unlike SSIM or LPIPS, CIS provides a domain-specific measure of ``culinary distance'' between images, making it suitable as a loss for generator training and for real-time cooking progress monitoring.

\section{Problem Formulation}
\label{sec:problem}

We aim to generate visually realistic cooked food images and monitor real-time cooking progress on edge devices. Our pipeline consists of three components:  
(1) an Edge-Efficient Text-guided Conditional Image Generator ($G_\theta$),  
(2) a Culinary Image Similarity (CIS) metric ($\mathcal{F}_{cul}$), and  
(3) a progress monitoring mechanism.  
The architecture is optimized for strict memory and latency budgets while remaining semantically expressive.  

\subsection{Edge-Efficient Text-guided Conditional Image Generator ($G_\theta$)}
\noindent\textbf{Inputs:}
\begin{itemize}[noitemsep,topsep=0pt]
    \item Raw food image $I_{\text{raw}} \in \mathbb{R}^{H \times W \times 3}$ captured by oven camera before cooking starts.  
    \item Recipe label $c \in \mathcal{C}$ (e.g., ``pizza'', ``chicken breast'', ``cheesecake''), obtained from a food recognition model or user input.  
    \item Cooking state $d_s \in \{{cs_1}, {cs_2},\dots {cs_n}\}$.  
\end{itemize}

The generator produces a target cooked image:
\begin{equation}
\hat{I}_{d_s} = G_\theta(I_{\text{raw}}, c, d_s).
\end{equation}

We adopt a U-Net architecture from \cite{10.5555/3495724.3496298} with encoder–bottleneck–decoder structure and skip connections, but modify it for text-guided conditional generation rather than iterative denoising. Naively scaling model capacity to capture food transformations is infeasible on edge devices, so we design two innovations:  
\begin{itemize}[noitemsep]
    \item \textbf{Text-guided Conditioned U-Net}, which injects recipe and cooking state context at every layer, and  
    \item \textbf{Specialized Composite Loss}, which combines adversarial, perceptual, and domain-specific supervision.  
\end{itemize}
For edge deployment, only the generator is required as it produces the final output. Thus, subsequent sections focus exclusively on our generator architecture innovations, while the discriminator remains unchanged as described in \cite{isola2017image}. Detailed architecture shown in Fig. \ref{fig:generatorarchitecture}.
\subsubsection{Text-guided Conditioned U-Net}
\begin{figure}
	\centering
	\includegraphics[width=0.9\linewidth]{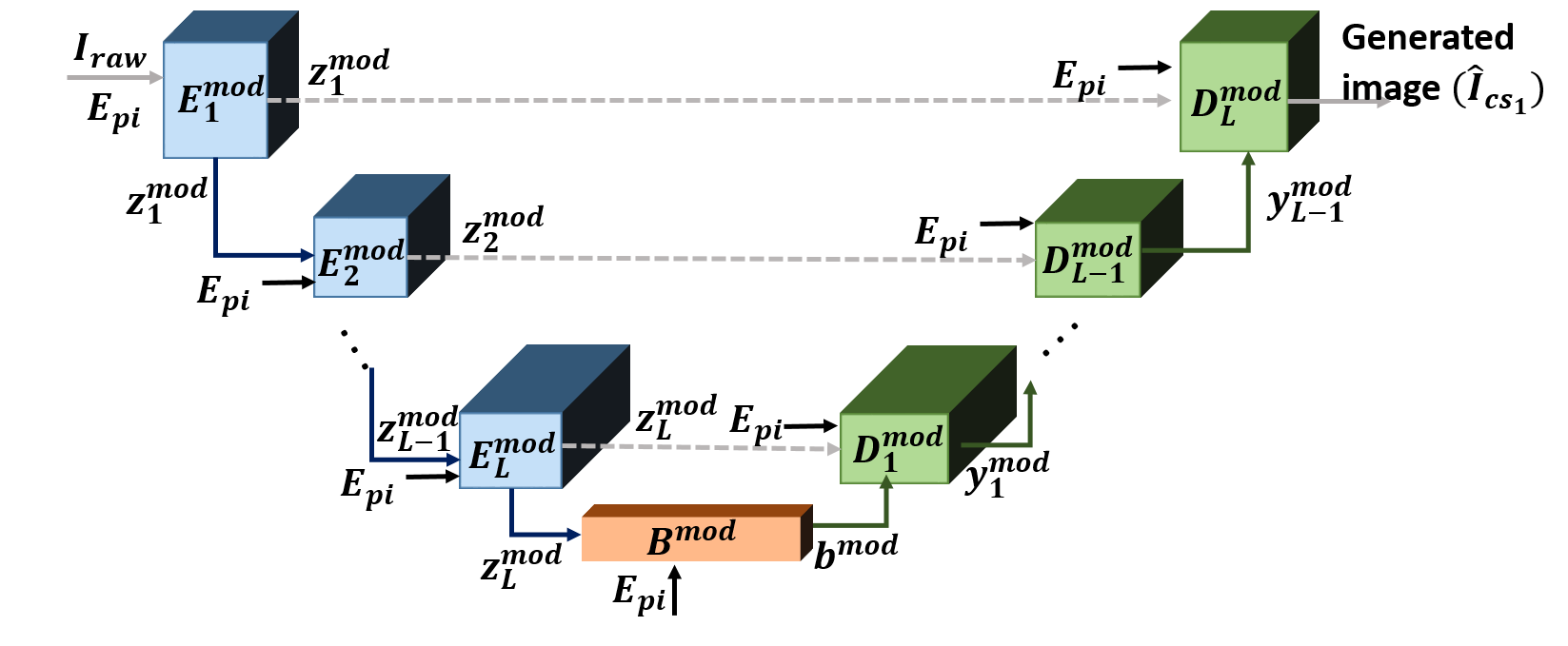}
	\vspace{0.1in} 
	\includegraphics[width=0.7\linewidth]{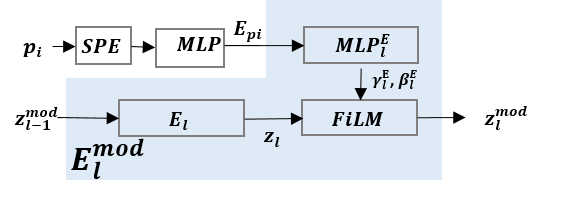}
	\caption{Text-guided Conditioned U-Net architecture. Top: Complete U-Net with feature modulation at each layer. Bottom: Detailed modulated layers. Encoder layers $E_l$ use contextual embeddings $E_{p_i}$ (recipe name, cooking state) for $FiLM$-based feature modulation. Decoder layers $D_l$ apply identical modulation for consistent context-aware processing }
	\vspace{-1em}
	\label{fig:unetwithmlp}
\end{figure}
Standard U-Nets treat translation as unconditioned, which is inefficient for food transformations. We guide the generator using context embeddings that modulate intermediate features (Fig.~\ref{fig:unetwithmlp}).

Each recipe-state pair $(c, d_s)$ is assigned an index $p_i$,  (e.g.,\texttt{(muffin,cs\textsubscript{1})}$\rightarrow$$p_1$,  \texttt{(muffin,cs\textsubscript{2})}$\rightarrow$$p_2$, \texttt{(fruitpie,cs\textsubscript{3})}$\rightarrow$$p_3$), which is then mapped to a learnable embedding using sinusoidal positional embedding $\mathcal{SPE}$ \cite{vaswani2017attention}:
\begin{equation}
E_{p_i} = \text{MLP}(\mathcal{SPE}(p_i)),
\end{equation} 
For each U-Net layer $l$, an MLP maps $E_{p_i}$ to scale and shift parameters:
\begin{equation}
[\gamma_l, \beta_l] = \text{MLP}_l(E_{p_i}),
\end{equation}
where $\gamma_l$ : Scale factor (controls the amplitude of features) for layer $l$ and  $\beta$ : Shift factor (adds bias to features) of layer $l$ and $\text{MLP}_l:\mathbb{R}^d \to \mathbb{R}^{d_l}$.

\noindent\textbf{Modulate Features}: The modulation parameters ($\gamma$ and $\beta$) are then injected at output ($z_l$ and $y_l$) of all intermediate layers of U-Net ($E_l$ and $D_l$) using Feature-wise Linear Modulation ($FiLM$) \cite{perez2017filmvisualreasoninggeneral}: 

\begin{equation}
\begin{aligned}
	&\textbf{FiLM:} \quad
	\mathrm{FiLM}(z;\gamma,\beta) = \gamma \odot z + \beta, \\[5pt]
	&\textbf{Encoder:} \\[2pt]
	&\quad z_l^{mod} = 
	\begin{cases} 
		I_{raw} & l = 0, \\ 
		\mathrm{FiLM}\!\big(E_l(z_{l-1}^{mod});\, \gamma_l^E, \beta_l^E\big) & l \in (0, L] 
	\end{cases} \\[8pt]
	&\textbf{Bottleneck:} \quad
	b^{mod} = \mathrm{FiLM}\!\big(B(z_L^{mod});\, \gamma^B, \beta^B\big), \\[5pt]
	&\textbf{Decoder:} \\[2pt]
	&\quad y_l^{mod} = 
	\begin{cases} 
		b^{mod} & l = 0, \\ 
		\mathrm{FiLM}\!\big(D_l(y_{l-1}^{mod}, z_{L-l}^{mod});\, \gamma_l^D, \beta_l^D\big) & l \in (0, L] 
	\end{cases}
\end{aligned}
\end{equation}
where $z_l^{mod}$ ,$y_l^{mod}$ are the modulated features of intermediate layer $l$ which goes as input to next layer. We call the combined unit $E_l^{mod}$ which combines $E_l$ and $FiLM$ as:
\begin{equation}
	E_l^{mod}(z_{l-1}^{mod},E_{p_i}) = \mathrm{FiLM}\!\big(E_l(z_{l-1}^{mod});\, \gamma_l^E, \beta_l^E\big)
\end{equation}

This mechanism allows to dynamically alter the statistical properties of the feature maps at every layer of the network. The context embedding does not merely concatenate with the image; it directly controls the internal state of the network's transformations, enabling precise, context-aware generation. This is the key to our model's efficiency, as it allows a single set of parameters to be dynamically modulated for a wide variety of culinary tasks.

\subsubsection{Specialized Composite Loss}
We train $G_\theta$ with a composite objective:
\begin{equation}
\mathcal{L}_{gen} = \lambda_1 \mathcal{L}_{GAN} + \lambda_2 \mathcal{L}_{LPIPS} + \lambda_3 \mathcal{L}_{CIS},
\end{equation} 
\begin{itemize}[noitemsep]
    \item $\mathcal{L}_{GAN}$: PatchGAN adversarial loss \cite{isola2017image}, enforcing local realism.  
    \item $\mathcal{L}_{LPIPS}$: perceptual similarity \cite{zhang2018unreasonable}, aligning with human judgment.  
    \item $\mathcal{L}_{CIS}$: our domain-specific loss, encouraging culinary plausibility and temporal consistency (defined in Sec.~\ref{sec:culinarysimilaritynet}).  
    \item $\lambda_1$, $\lambda_2$, $\lambda_3$ are hyperparameters to control contribution of each loss
\end{itemize}
Let $I_{d_s}$ be the real image from cooking session at cooking state $d_s$ and $\hat{I}_{d_s}$ be the generated images of this state. We define $\mathcal{L}_{CIS}$:
\begin{equation}
	\mathcal{L}_{CIS} = 1 - \mathcal{F}_{cul}(I_{d_s}, \hat{I}_{d_s})
	\label{eq:cisloss}
\end{equation}
Training details provided in Appendix \ref{appendix:experimentgen}
\begin{figure}
	\centering
	\includegraphics[width=0.99\linewidth]{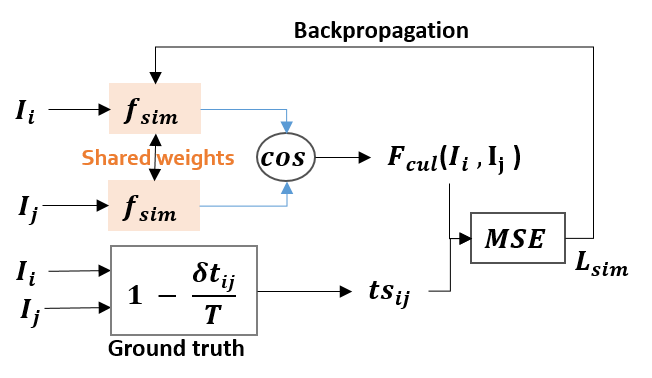}
	\caption[]{We learn a culinary image similarity metric, $\mathcal{F}_{cul}$, by leveraging temporal distances between image pairs from cooking sessions. The Siamese network, $f_{sim}$, is trained to map cooking stages into an embedding space, where temporal progression is represented as a smooth, continuous trajectory.}
	\label{fig:matchingmodel}
\end{figure}
\subsection{Culinary Image Similarity (CIS) Metric}
\label{sec:culinarysimilaritynet}

Generic metrics like SSIM and LPIPS correlate poorly with cooking dynamics, as they fail to capture gradual, domain-specific changes (e.g., subtle browning, dough rise, size change). We propose CIS, a similarity metric tailored to cooking transformations.  

Given two images $I_i$ and $I_j$ from a cooking session, a Siamese network $f_{sim}$ embeds them into a space where temporal progression forms a smooth trajectory. Similarity is measured as:
\begin{equation}
\mathcal{F}_{cul}(I_i, I_j) = \cos(f_{sim}(I_i), f_{sim}(I_j)), \quad \mathcal{F}_{cul} \in [0,1].
\end{equation}
We call this learned metric as Culinary Image Similarity $(CIS)$ Metric. This metric is specifically used to find similarity between two images from the ``same'' cooking session at different timesteps. This metric is also used as a loss function to train the generator and yield temporally consistent generated cooked images. It is also used for cooking progress monitoring where the similarity value between two images (generated and real) can guide how far the cooking has progressed in terms of time.

To achieve this, we first train a specialist Siamese network, $f_{sim}$ as shown in Fig. \ref{fig:matchingmodel}, from scratch to act as an ``expert" on the process of cooking. This network learns an embedding space where the progression of a cooking session is represented as a smooth, continuous trajectory. 

We construct a dataset of cooking sessions where each session consists of a sequence of images of a recipe captured at a frequency of 30 secs. We assign a temporal score  $ts_{i,j}$ to each frame pair based on their separation in time. Specifically, $ts_{i,j} = 1 -  \frac{\delta t_{i,j}}{T}$, where $\delta t_{i,j}$ is the time separation between two frame $i$ and $j$ and $T$ is the length of cooking session. The distance between 1st frame and last fame is marked as max i.e. 1 and the distance between same fame is considered as 0. Other distances are scaled between 0 and 1 based on temporal distance.  The objective is to learn a function $f_{sim}$ such that the cosine similarity between embedding’s reflects the culinary proximity of the input images. We train a Siamese network on this dataset using a contrastive loss between image pairs. Since we have continuous labels $\in [0,1]$, we use MSE loss to train the model. 

Training pairs are labeled with this temporal proximity score $ts_{i,j}$, and $f_{sim}$ is optimized with MSE:
\begin{equation}
\mathcal{L}_{sim} = \text{MSE}(\mathcal{F}_{cul}(I_i, I_j), ts_{i,j}).
\end{equation}

CIS thus provides a dynamic, domain-aware measure of ``culinary distance.'' It is used both as a loss for generator training (Eq.~\ref{eq:cisloss}) and as a signal for progress monitoring.  
Training details provided in Appendix \ref{appendix:experimentCIS}
\subsection{Cooking Progress Monitoring}

At inference, the system generates $n$ images of different cooking states ($cs_1, cs_2,\dots, cs_n$). The user selects the desired endpoint  $\hat{I}_{cs_d}$ reflecting personal doneness level.  

During cooking, each observed frame $I_t$ is compared to the chosen target $\hat{I}_{cs_d}$ using CIS:
\begin{equation}
\text{progress}(t) = \mathcal{F}_{cul}(I_t, \hat{I}_{cs_d}).
\end{equation}
Cooking stops when peak similarity is detected within a moving window, indicating that the real dish visually matches the user selected generated image. This decision requires only the camera feed and user chosen generated image, with no additional sensors or external probes.

\section{Results and Analysis}
\label{sec:results}
\subsection{Cooked Food Image Generation}
We now present a comprehensive evaluation of our proposed method for image generation. We demonstrate that our model achieves better results in culinary image-to-image translation as compared to baselines, while being significantly more efficient than existing methods.
\subsubsection{Dataset}
\label{sec:datasetimagegen}
In the cooking domain, no existing dataset comprehensively captures the culinary process across various recipes. YouCook2 \cite{ZhXuCoAAAI18} provides detailed annotations of cooking activities like "stirring" and "adding salt," while COM Kitchens \cite{maeda2024comkitchensuneditedoverheadview} offers overhead-view cooking videos annotated with visual action graphs linking recipe instructions to visual elements. However, these datasets focus on cooktop activities and lack monitoring of food item transitions, such as color and texture changes, during cooking. To address this gap, we introduce a robust oven-based cooking progression dataset, collected in controlled settings with chef-annotated cooking states. The dataset includes 1708 cooking sessions across 30 diverse recipes, covering bakery and meat items, providing a comprehensive view of cooking state progression.

\textbf{Data Collection}: Cooking progress images were obtained using a an AI Oven equipped with a top-mounted stationary camera that recorded frames at 30 sec intervals. Each session involved a single food item, supported by the food recognition model, placed clearly on a plate and cooked without opening the oven door until at least one edible state was reached. To enhance dataset diversity, multiple versions of the same dish were prepared (e.g., pizzas with varied toppings, salmon in different portion sizes). Cooking was done using oven’s auto mode to ensure uniform temperature, timing and cooking mode setting across all sessions of each recipe. 

The collected sessions were annotated for three visually distinct edible states for each food item.
 
The first image of the session, where the food is clearly visible, is designated as ``\textit{raw}" image. Subsequently, the three cooking states are designated as \textit{``basic''} cook, \textit{``standard''} cook, and \textit{``extended''} cook based on the degree of cooking. \textit{``basic''} cook refers to the minimal cooking state where the food item is rendered edible but might lack additional texture, \textit{``standard''} cook represents the ideal cooking state as per the established guidelines for a particular recipe and \textit{``extended''} cook denotes a state where the food item is subjected to prolonged cooking to achieve enhanced characteristics such as browning, crispiness, or caramelization. Some samples images of collected sessions are shown in Figure \ref{fig:session} in Appendix \ref{appendix:dataset}. Going forward in the results section, we will deal with three cooking states only and will refer $cs_1, cs_2, cs_3$ as \textit{``basic''}, \textit{``standard''} and \textit{``extended''} respectively. Our proposed model is not constrained by a fixed number of cooking states, allowing for varying cooking state counts per recipe (e.g., 2 states for oven chips and 4 states for pizza). For uniformity in our experiments, we standardized all recipes to 3 cooking states.

 A total of 1,708 sessions for 30 recipes were collected and annotated. Complete list of recipes provided in Table \ref{tab:recipes} in Appendix \ref{appendix:dataset}. The dataset was divided into (train,val,test) set in the ratio of 70:10:20. To enhance variability during the training phase, we applied various geometric augmentations, including flipping and rotation. 

\subsubsection{Baseline and Metric}
We benchmark our proposed model against two prominent image-to-image translation frameworks: Pix2Pix \cite{isola2017image} and Pix2Pix-Turbo \cite{parmar2024onestepimagetranslationtexttoimage}. Pix2Pix, while foundational in the domain of image translation, does not inherently support textual inputs. Consequently, we trained three separate models for the distinct cooking states. Pix2Pix-Turbo was selected as the baseline due to its capability to accept both textual prompts and image conditioning, aligning closely with our experimental setup. For instance, textual prompts such as ``extra cooked Croissants in oven'' are utilized to guide the generation process. 

We evaluate our model using Fréchet Inception Distance (FID) \cite{heusel2018ganstrainedtimescaleupdate} and Learned Perceptual Image Patch Similarity (LPIPS) \cite{zhang2018unreasonable}. Lower scores are better for both metrics. 
\subsubsection{Results} 
\begin{table}[]
\centering
\caption{Comparison with state-of-the-art methods and ablation
	of key components on our curated dataset. Our model achieves best performance while requiring significantly fewer parameters .
	(Only generator parameters are mentioned. FID and LPIPS calculated on 715 image pairs)}
\resizebox{\columnwidth}{!}{%
\begin{tabular}{lrrrr}
\toprule
Method & \multicolumn{2}{c}{Parameters (M)} & FID\( \downarrow \) & LPIPS\( \downarrow \) \\
\cmidrule(lr){2-3}
& Trainable  & Total  & & \\
\midrule
Pix2Pix  & 163.23\footnote  & 163.23 & 153.00 & 0.4711 \\ 
Pix2Pix-Turbo & 9.00 & 1290.00 & 75.42 & 0.2523 \\
Ours & \textbf{8.68} & \textbf{8.68} & \textbf{52.18} & \textbf{0.2145} \\
\midrule
\midrule
w/o Recipe Guidance & 8.68 & 8.68 & 58.74 & 0.2398 \\
w/o $\mathcal{L}_{CIS}$ & 8.68 & 8.68 & 54.98 & 0.2310 \\
\bottomrule
\end{tabular}%
}
\label{tab:pix2pix}
\end{table}
\footnotetext{The number of parameters for ``Pix2Pix" is calculated based on three separate models for each cooking state.}

Our model was comprehensively evaluated across 30 diverse recipes and 346 cooking sessions, comprising 346 \textit{raw-basic}, 285 \textit{raw-standard}, and 84 \textit{raw-extended} pairs, making total test set of 715 image pairs. As shown in Table~\ref{tab:pix2pix}, it consistently outperforms both Pix2Pix and Pix2Pix-Turbo, achieving the lowest FID (52.18) and LPIPS (0.2145) scores while requiring only 8.68M parameters - nearly 19× fewer than Pix2Pix - highlighting its suitability for efficient deployment.  Unlike Pix2Pix, which relies solely on paired image-to-image translation, our framework integrates text-guided conditioning that encodes recipe-specific context. This additional semantic guidance enables the model to capture fine-grained variations in ingredient appearance, shape, and cooking state, leading to more contextually relevant outputs. In addition, the CIS loss explicitly enforces perceptual similarity between generated outputs and ground-truth images by incorporating domain-specific temporal cues. In contrast, Pix2Pix-Turbo, although generating visually coherent images, often fails to represent critical domain changes such as ingredient size change, browning, or texture shifts

\textbf{Ablation}: To further validate the contribution of these components, we conducted ablation studies. Removing recipe guidance (by removing recipe name from $FiLM$-based feature modulation input) degraded performance as the model did not had any recipe specific semantic cues. Hence it was not able to learn the complex transformations that are specific to each recipe, e.g. meat items shrink in size after cooking while cookies expand in size. Removing the CIS loss reduced the model to adversarial and perceptual training, leading to inferior texture and color reproduction and less visual difference in three generated cooking states. Both studies highlight the complementary roles of recipe guidance and CIS loss in driving the observed improvements. Collectively, these results demonstrate that our approach not only surpasses existing baselines in accuracy and perceptual quality but also achieves superior efficiency, making it highly practical for real-world culinary image translation tasks.

\begin{figure}[h]
	\centering
	\includegraphics[width=0.99\linewidth]{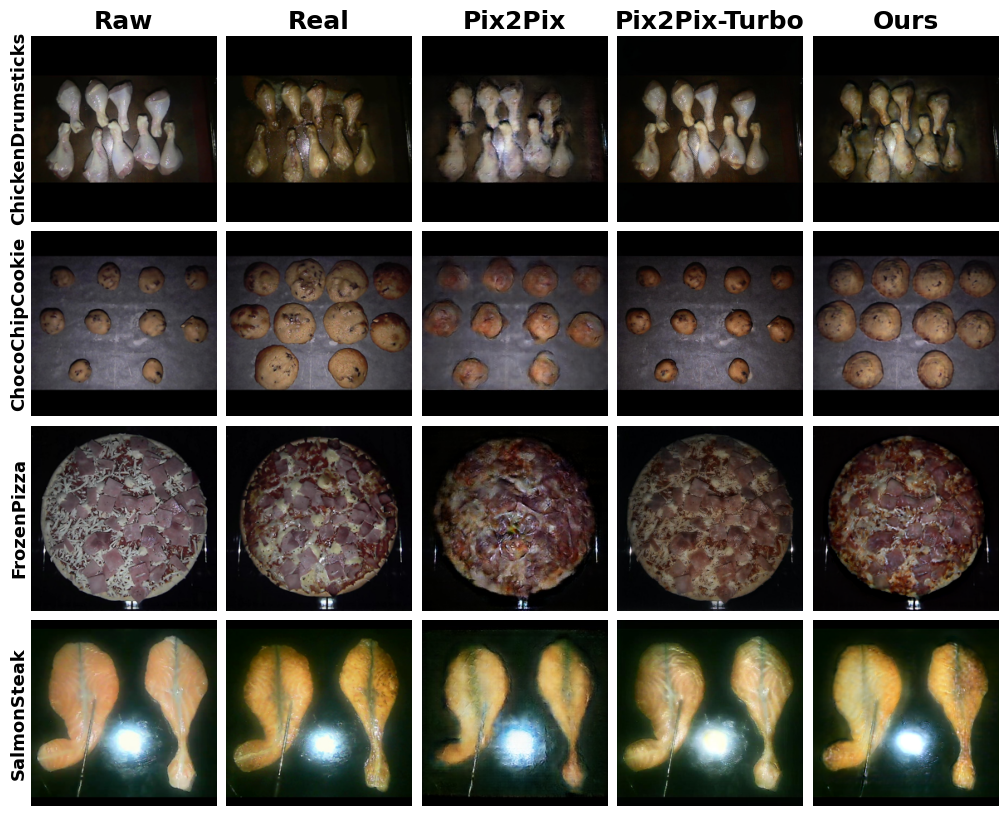}
	\caption{Generated image quality comparison across baseline methods and our proposed approach. The results demonstrate the performance of all models for a specific cooking state, illustrating that our proposed method produces the most plausible images. (Best viewed in color)}
	\label{fig:results}
\end{figure}

\textbf{Generated Image Quality:} 
We present qualitative results to visually demonstrate the superiority of our method. Figure \ref{fig:results} shows a grid of images. Each row correspond to a different recipe and column shows Raw, Real, Pix2Pix generated, Pix2Pix-Turbo generated and our method generated images of a cooked state. The visual comparison clearly shows that our method generates sharper textures, more accurate colors (e.g., better browning on the ChickenDrumStick, increase in size of ChocoChipCookie, clear edges in SalmonSteak), and more realistic structural changes compared to the baselines, which often produce blurrier or less plausible results.

Figure \ref{fig:teaser} also demonstrates the controllability of our model for generating multiple cooking state images from raw image and cooking state as textual input. Each row feature a single recipe and its raw, three cooking state generated images compared with real image from the session. Each generated image has visually distinct color and texture. This proves that the recipe and cooking state guidance allows for fine-grained control over the cooking process, generating a plausible and consistent progression of cooking transformations.

\begin{figure}[h]
	\centering
	\includegraphics[width=\linewidth]{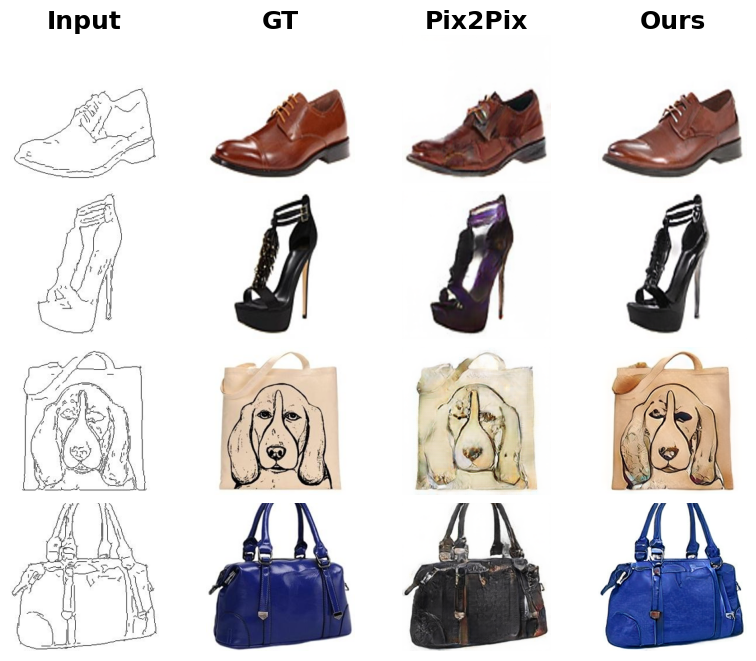}
	\caption{Experimental results on public datasets. The top two images showcase results from the edge2shoes dataset, while the bottom two images demonstrate performance on the edge2handbags dataset. Our proposed model generates images that exhibit higher fidelity to the ground truth. (Best viewed in color)}
	\label{fig:edge}
\end{figure}

\textbf{Comparison with public dataset}: We also conduct a comparative analysis between our proposed model and the original Pix2Pix framework on two widely recognized datasets: edge2shoes \cite{6909426} and edge2handbags \cite{zhu2018generativevisualmanipulationnatural}. The datasets are utilized with the train and validation splits as provided in \cite{isola2017image}. To establish a robust evaluation protocol, we randomly sample 10\% of the image pairs from the training data to create a dedicated test set. For model selection, we identify the checkpoint yielding the best FID score on the validation set and report the corresponding test results for both models.
 
In this experiment, for our proposed model, we removed the recipe name and cooking state text guidance, as all images belong to the same class. Additionally, we omitted the $\mathcal{L}_{CIS}$ loss term due to the absence of temporal dynamics in these datasets. Our experimental findings, summarized in Table \ref{tab:resultspublicdata}, demonstrate that our proposed model consistently outperforms the original Pix2Pix framework across both datasets. This improvement can be attributed to our enhanced loss function, which replaces the L1 loss in Pix2Pix with LPIPS and architectural optimizations in generator U-Net that comes from \cite{10.5555/3495724.3496298}. To further validate the versatility of our approach, we combined the edge2shoes and edge2handbags datasets into a unified dataset. In this combined setting, we employed the dataset name as a text prompt to our model, effectively showcasing its ability to train a single model capable of handling multiple classes. Examples of generated images for edges2shoes and edges2handbags are provided in Figure \ref{fig:edge}. Pix2Pix-Turbo did not yield satisfactory results when using a single prompt for the entire dataset, and manually tailoring prompts for each sample was not feasible. Hence, the results are not reported here, though this direction may be further explored in future work.

\begin{table}[h]
	\centering
	\caption{Quantitative comparison of FID and LPIPS scores on validation and test sets for Pix2Pix and our proposed method using public datasets. Combined results are omitted for Pix2Pix due to its lack of text input support. Edge2shoes val set has 200 and test set has $\approx$5k image pairs; edge2handbags val set has 200 and test set has 10k image pairs.}
	\label{tab:resultspublicdata}
	\resizebox{\linewidth}{!}{
		\begin{tabular}{lrrrrrrrr}
			\toprule
			\multirow{3}{*}{Dataset} & \multicolumn{4}{c}{Pix2Pix} & \multicolumn{4}{c}{Ours} \\ 
			\cmidrule(lr){2-5} \cmidrule(lr){6-9}
			& \multicolumn{2}{c}{FID $\downarrow$} & \multicolumn{2}{c}{LPIPS $\downarrow$} & \multicolumn{2}{c}{FID $\downarrow$} & \multicolumn{2}{c}{LPIPS $\downarrow$} \\ \cmidrule(lr){2-3} \cmidrule(lr){4-5}\cmidrule(lr){6-7} \cmidrule(lr){8-9}
			& Val & Test & Val & Test & Val & Test & Val & Test \\
			\midrule
			Edge2shoes & 57.89 & 23.28 & 0.217 & 0.217 & 34.50 & \textbf{7.40} & 0.166  & \textbf{0.163} \\
			Edges2handbags & 77.99 & 25.34 & 0.305 & 0.330 & 56.72 & \textbf{9.32} & 0.236 & \textbf{0.243} \\
			Combined & - & -& - & - & 42.43& 10.51 & 0.210 & 0.227 \\
			\bottomrule
		\end{tabular}
	}
\end{table}
We also assessed the Gemini API for few shot generation of cooked food images. Despite extensive prompt engineering, the model was unable to generate realistic images that reliably distinguished the three cooking target states, highlighting its current limitations for fine-grained food state representation. Detailed results are provided in Appendix \ref{appendix:experimentgen}.

\subsection{Culinary Image Similarity Metric}
We present a comprehensive evaluation of our proposed cooking similarity metric, demonstrating its ability to capture intricate cooking transformations, that traditional metrics like LPIPS and SSIM fail to address.
\subsubsection{Dataset and Metric}
We trained our proposed similarity metric using the same curated dataset described in Section \ref{sec:datasetimagegen}. As detailed in Section \ref{sec:culinarysimilaritynet}, our approach eliminates the need for manual annotations, leveraging temporal scores assigned to images sampled at equal intervals from each cooking session. To enhance the model's robustness, we introduced random rotations of sampled images during training, compelling the model to prioritize color and texture features for accurate similarity assessment.

We compared our proposed metric against standard metrics such as LPIPS \cite{zhang2018unreasonable} and SSIM \cite{wang2004image} as they are widely used in image similarity tasks. LPIPS relies on high-level perceptual features and SSIM emphasizes structural similarity on pixel level, and both of them fails to capture temporal evolution and task-specific nuances like browning or texture changes.  Our method, by contrast, is specifically designed to address these nuanced transformations.

\subsubsection{Results}
\begin{figure}[]
	\centering
	\begin{minipage}[b]{0.48\columnwidth}
		\includegraphics[width=\textwidth]{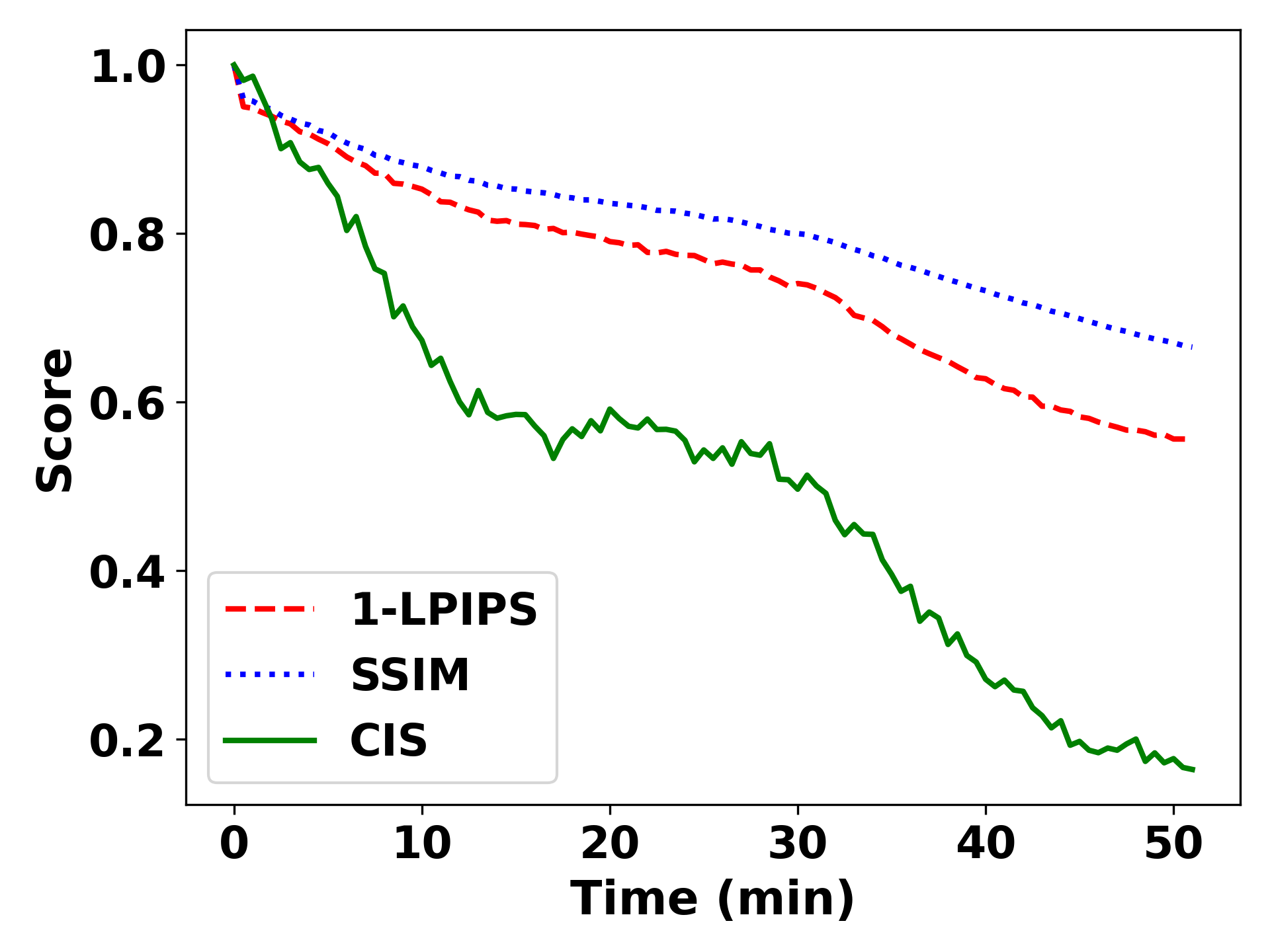}
		\label{fig:image1}
	\end{minipage}
	\hfill
	\begin{minipage}[b]{0.48\columnwidth}
		\includegraphics[width=\textwidth]{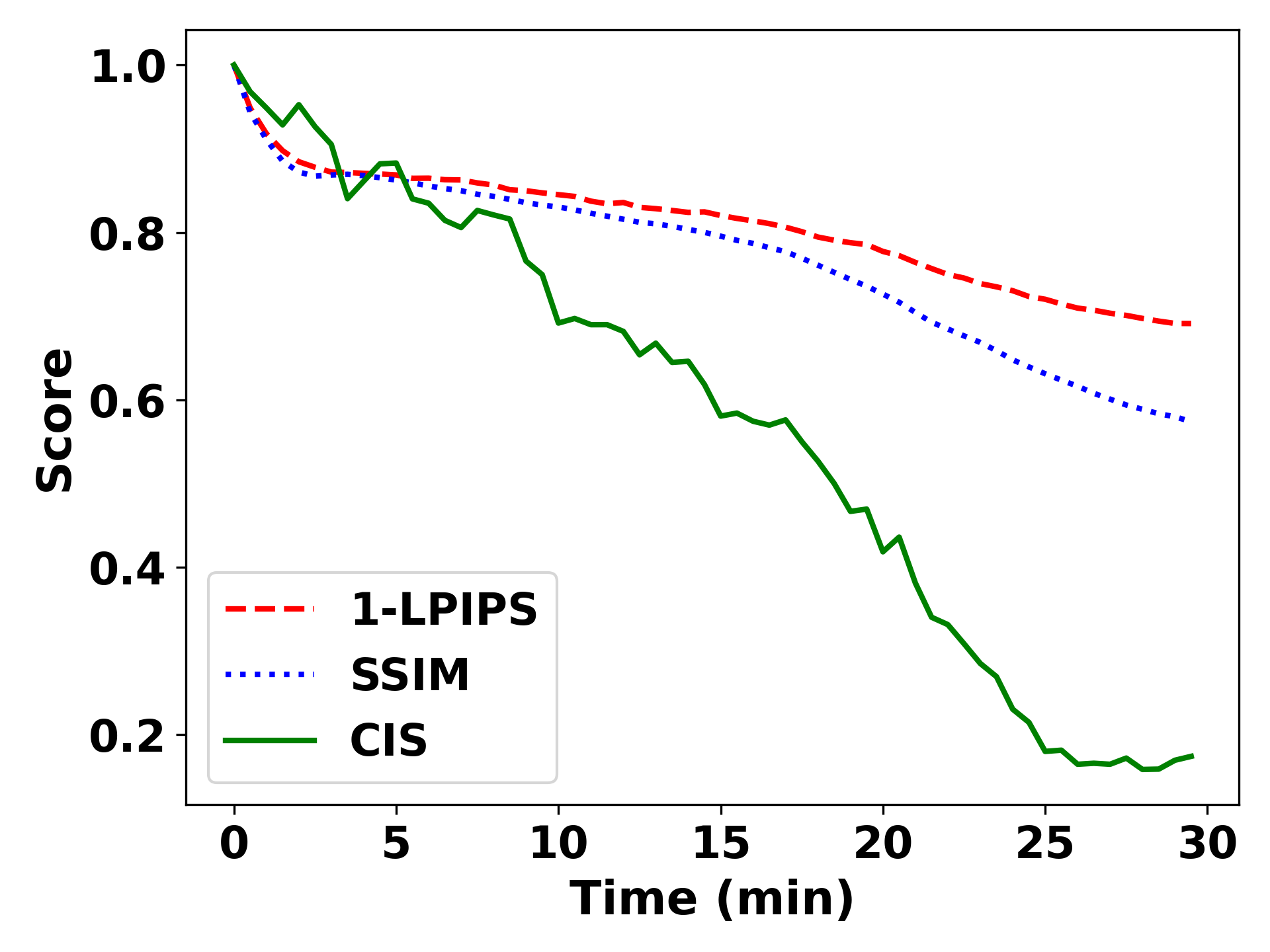}
		\label{fig:image2}
	\end{minipage}
	
	\vspace{-1em}
	
	\begin{minipage}[b]{0.48\columnwidth}
		\includegraphics[width=\textwidth]{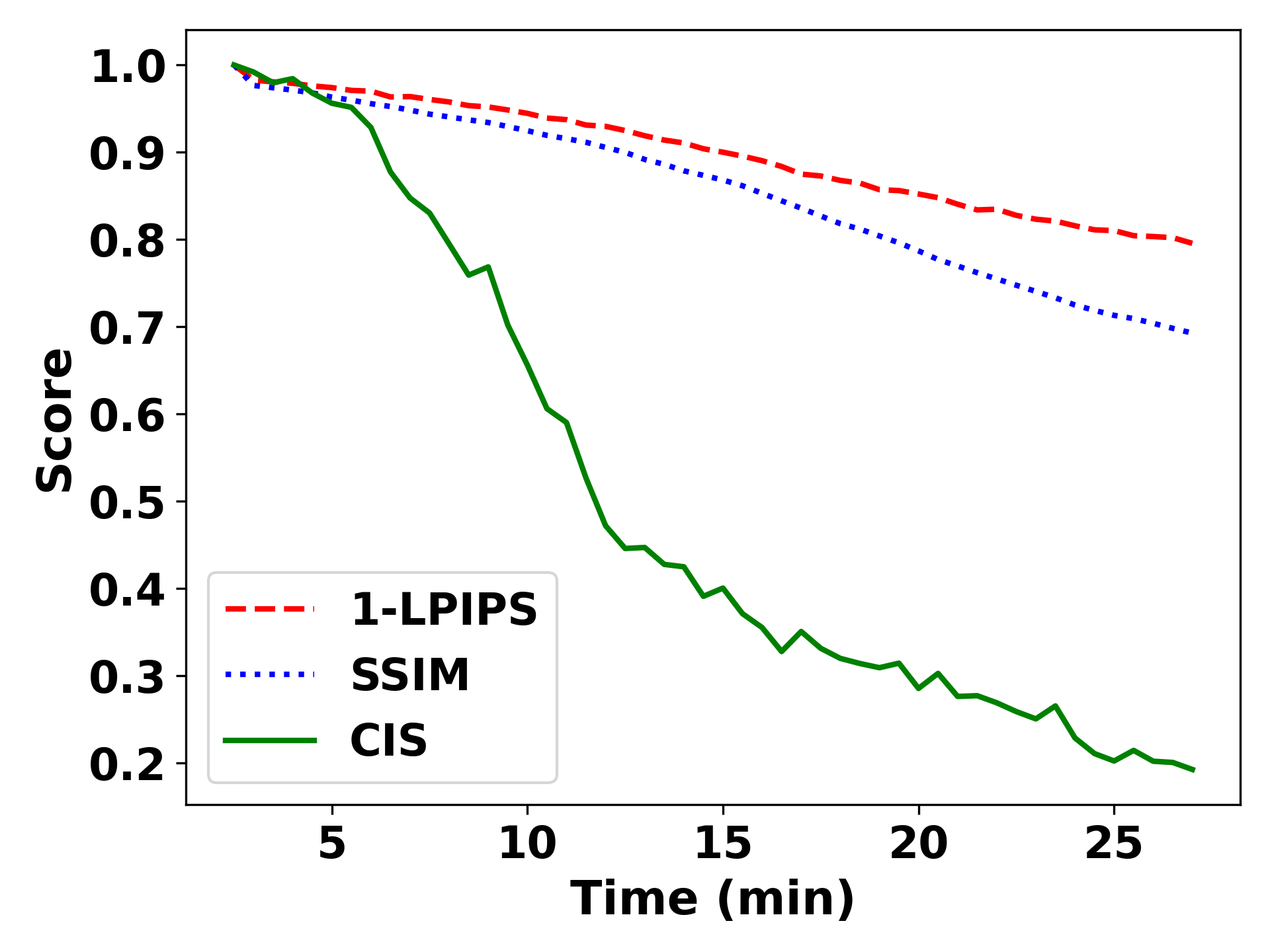}
		\label{fig:image3}
	\end{minipage}
	\hfill
	\begin{minipage}[b]{0.48\columnwidth}
		\includegraphics[width=\textwidth]{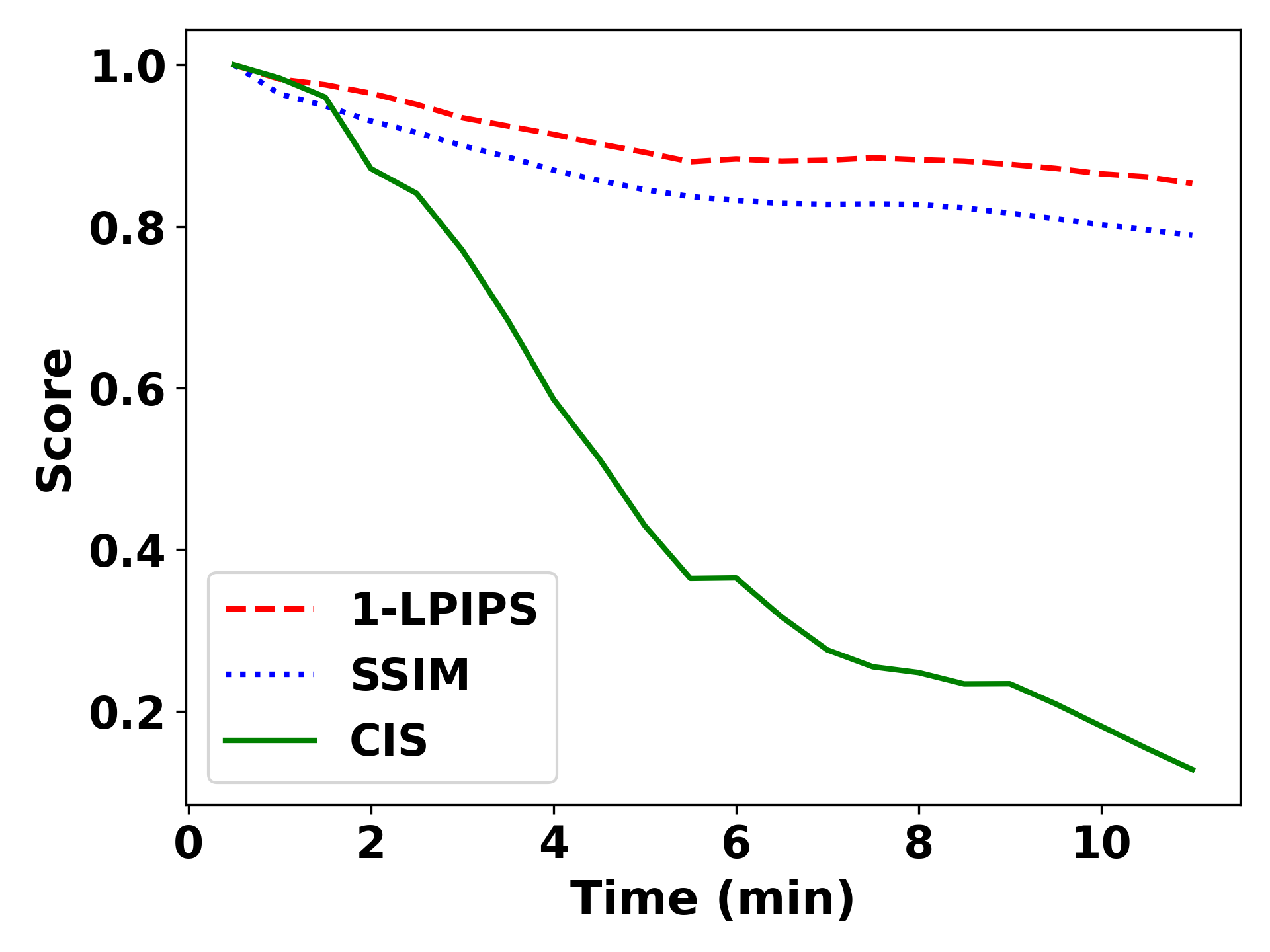}
		\label{fig:image4}
	\end{minipage}
	\vspace{-2em}
	\caption{Raw image similarity scores across session data using CIS, 1-LPIPS, and SSIM metrics for 4 different recipes. The CIS metric demonstrates variation ranging from 1.0 to $\approx$0.1, whereas LPIPS and SSIM exhibit a considerably narrower range of values.}
	\label{fig:gridreal}
\end{figure}

\begin{figure}[]
	\centering
	\begin{minipage}[b]{0.48\columnwidth}
		\includegraphics[width=\textwidth]{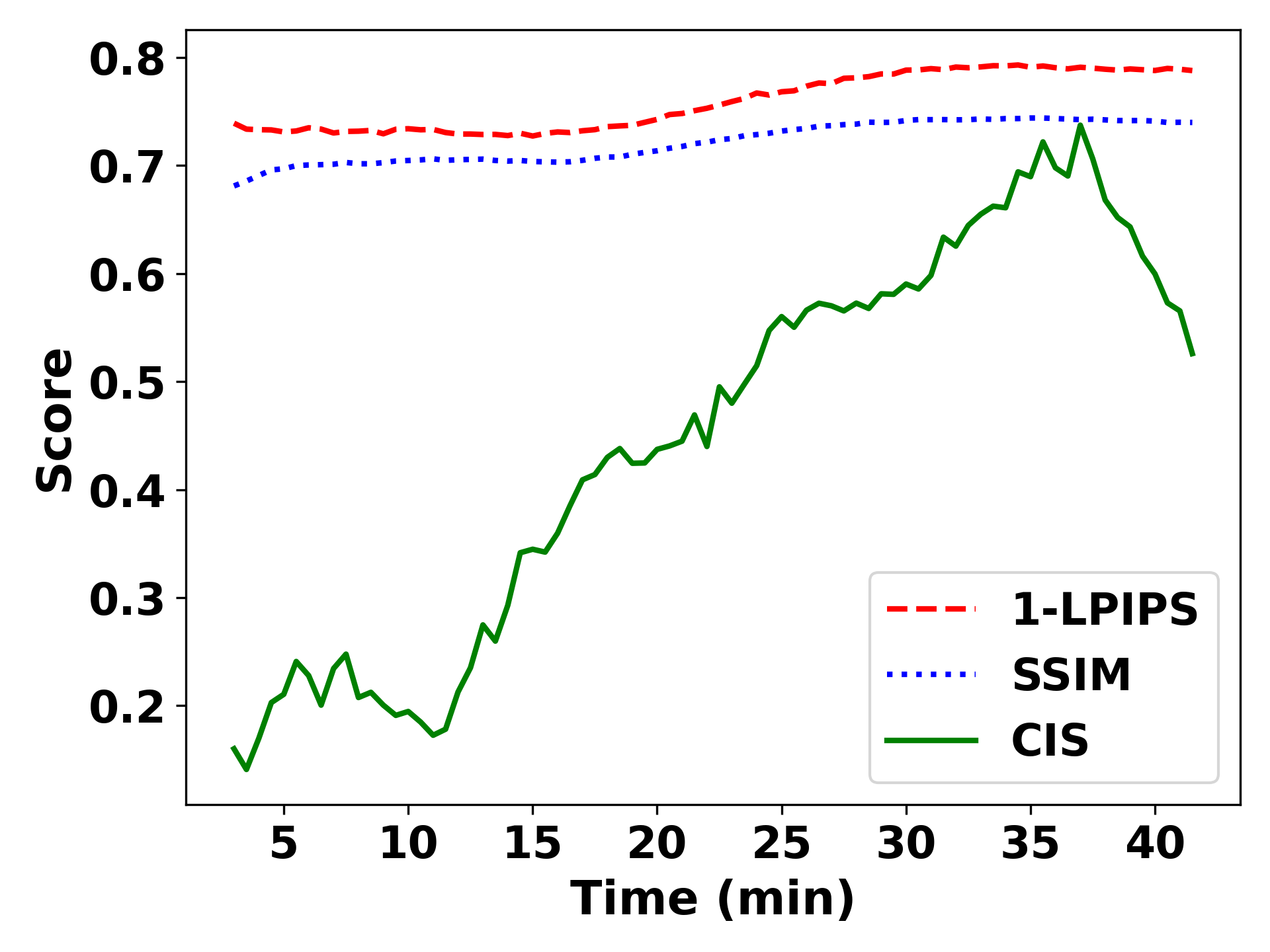}
		\label{fig:image1}
	\end{minipage}
	\hfill
	\begin{minipage}[b]{0.48\columnwidth}
		\includegraphics[width=\textwidth]{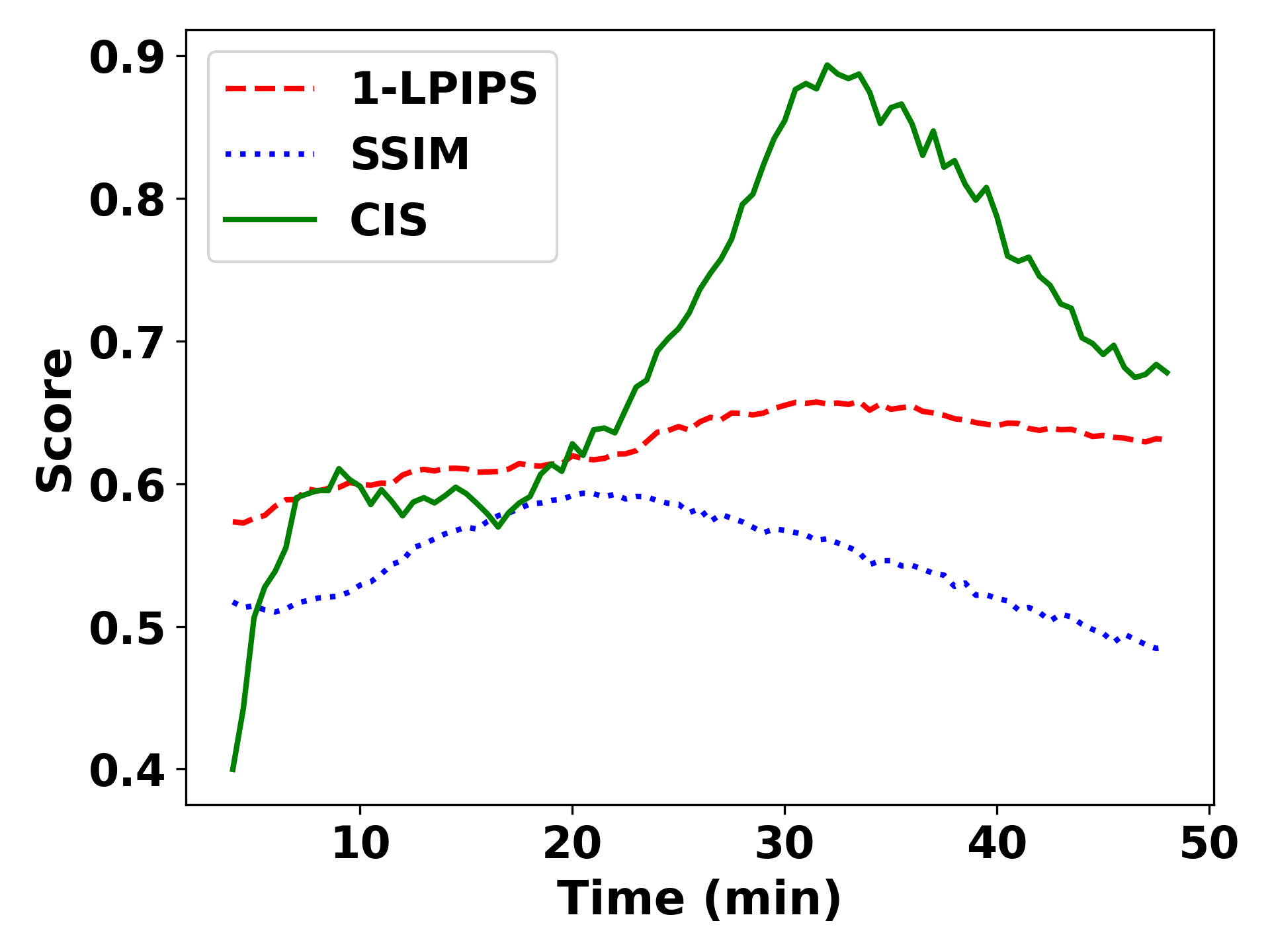}
		\label{fig:image2}
	\end{minipage}
	\vspace{-1em}
	
	\begin{minipage}[b]{0.48\columnwidth}
		\includegraphics[width=\textwidth]{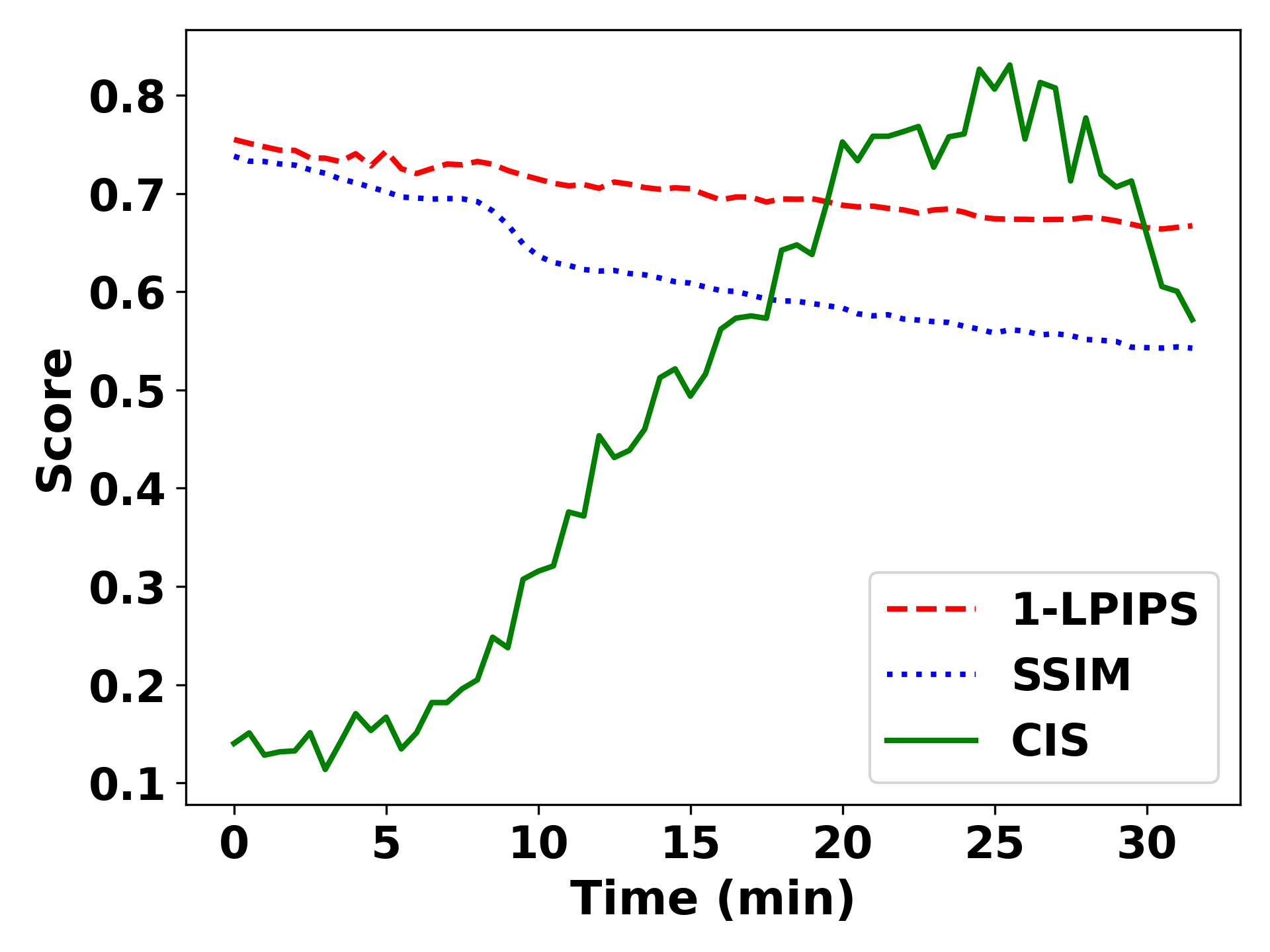}
		\label{fig:image3}
	\end{minipage}
	\hfill
	\begin{minipage}[b]{0.48\columnwidth}
		\includegraphics[width=\textwidth]{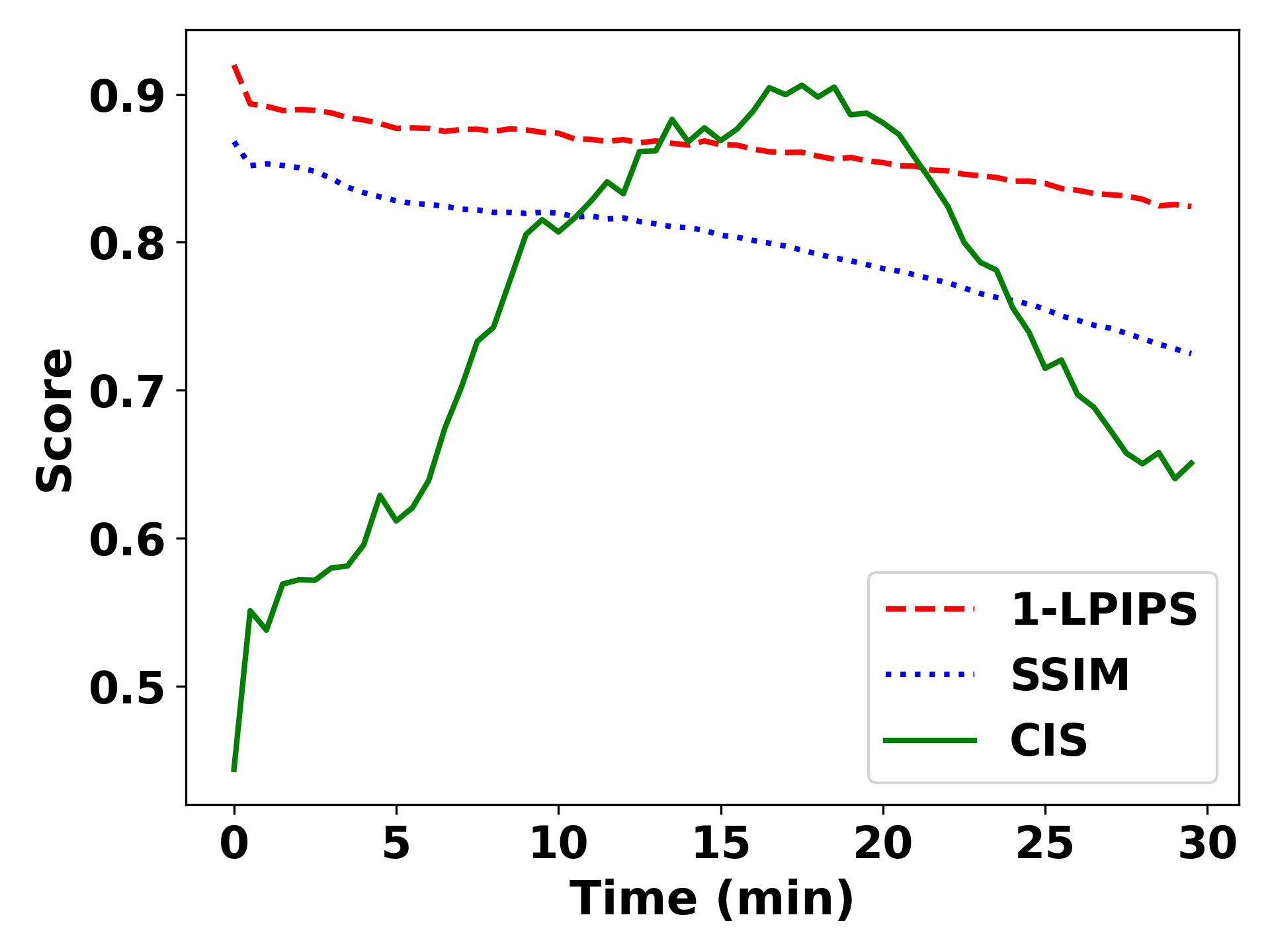}
		\label{fig:image4}
	\end{minipage}
	\vspace{-2em}
	\caption{Generated image similarity scores across session using CIS, 1-LPIPS and SSIM for 4 different recipes. Peak similarity value is used as cooking stopping criteria. More details in Appendix \ref{appendix:cpm}.}
	\label{fig:gridfake}
\end{figure}
To assess the performance of our proposed metric, we conducted an experiment comparing SSIM, LPIPS, and CIS scores across session images. We calculated all metrics for the following image pairs: (Raw-Raw), (Raw-Basic), (Raw-Standard), and (Raw-Extended), with the results presented in Table \ref{tab:metrics}.
Our analysis revealed that the CIS metric exhibits a significant range of variation, from 1.0 to 0.214, effectively distinguishing between session images captured at different times. In contrast, LPIPS and SSIM show much smaller ranges of variation—0.0 to 0.26 and 1.0 to 0.628, respectively—highlighting their limited ability to detect subtle visual differences during cooking.

\begin{table}[h!]
	\centering
	\caption{Comparison of SSIM, LPIPS and CIS for different cooking states averaged across test set.}
	\resizebox{0.7\columnwidth}{!}{%
		\begin{tabular}{lS[table-format=1.4]S[table-format=1.4]S[table-format=1.4]}
			\toprule
			\textbf{Pair} & \textbf{SSIM}$\uparrow$ & \textbf{LPIPS}$\downarrow$ & \textbf{CIS}$\uparrow$ \\
			\midrule
			Raw vs Raw & 1.0 & 0.0 & \textbf{1.0} \\
			Raw vs Basic & 0.722 & 0.209 & \textbf{0.519} \\
			Raw vs Standard & 0.663 & 0.235 & \textbf{0.366} \\
			Raw vs Extended & 0.628 & 0.260 & \textbf{0.214} \\
			\bottomrule
		\end{tabular}
	}
	\label{tab:metrics}
\end{table}

In Figure~\ref{fig:gridreal}, we use the first image of the session (raw image) as an anchor and compare it to all subsequent images in the session. CIS exhibits a wide dynamic range, from 1.0 to $\approx$0.1, whereas (1-LPIPS) and SSIM remain within a narrow band (1.0 to $\approx$0.7). This further supports the ability of our metric to detect small but meaningful changes that occur during the process, while traditional metrics fail to capture such variations. The CIS graphs exhibit a non-linear trend, characterized by varying slopes that reflect the dynamic nature of cooking processes. Portions with higher slopes indicate rapid changes in color and texture, while sections with smaller slopes correspond to slower transformations during cooking. This non-linear behavior highlights the metric's ability to capture the complex and evolving visual characteristics of food as it cooks.

In Figure~\ref{fig:gridfake}, we repeat the analysis using a generated image of a cooking state as the anchor and compare it across the session. CIS clearly identifies a peak similarity between the generated and real images. Maximum similarity value for CIS varies from 0.6 (for complex recipes like pizza) to 0.9 (for simple recipes like cheesecake), with an average of 0.7 on the test set. Using a sliding window approach, we locate this peak where maximum similarity occurs to determine the optimal stopping point during cooking when a user selected cooked state has reached. However, SSIM and LPIPS again show limited variation and no discernible peak, indicating that these metrics are not suitable for identifying meaningful stopping criteria, even when thresholds are manually adjusted. The CIS metric has been successfully deployed on-device for cooking progress monitoring, achieving a processing time of 0.3 seconds per comparison between generated and real images. In human evaluations, the system demonstrated robust performance, accurately stopping cooking at the user desired doneness in $\approx$90\% of cases.

\section{Conclusion} 
We introduced an edge-efficient generative framework and a new cooking-progression dataset to model visual food transformations inside an oven. Our method synthesizes target cooked appearances based on raw input, recipe, and desired doneness, enabling user-preferred outcomes instead of fixed time-based controls. The Culinary Image Similarity (CIS) metric supervises generation and provides a visual stopping signal, forming a closed-loop system that mimics human cooking behavior. Unlike regression or classification approaches, our generation-based formulation offers instance-specific guidance and interpretable visual feedback. The compact conditional U-Net design runs in real time on a consumer-grade NPU, enabling practical on-device deployment for smart kitchen appliances.

Our system currently uses data collected with the oven door closed and from a single top-down camera. Real-world scenarios involving door openings (changing light conditions) or food flipping were not explicitly modeled. While rotation and flipping augmentation were incorporated during training, the model's robustness to in-session food movement or varying lighting conditions remains untested. Future work will explore these scenarios to enhance generalization.

Beyond oven control, the chef-annotated dataset and progression modeling approach lay the groundwork for multimodal cooking intelligence, including doneness assessment, food-state tutoring, and visual explanations of culinary transformations. This framework also suggests a pathway for improving user experiences across home devices by incorporating preference-driven, visually grounded interactions over static heuristics, offering a transferable paradigm for domains where appearance evolves over time.

{\small
\bibliographystyle{ieee_fullname}
\bibliography{egbib}
}

\clearpage
\appendix
\section{Dataset}
\label{appendix:dataset}
We collected total 1708 cooking sessions for 30 recipes. Details of each recipe are provided in \ref{tab:recipes}.
\begin{table}[h]
	\centering
	\small 
	\rowcolors{2}{lightgray}{white}
	\begin{adjustbox}{max width=\columnwidth}
		\begin{tabular}{c>{\raggedright\arraybackslash}p{3.2cm}c}
			\toprule
			\textbf{S. No.} & \textbf{Recipe Name} & \textbf{Sessions} \\
			\midrule
			1 & Baguettes & 25 \\
			2 & BellPepperChunks & 24 \\
			3 & BrusselSprouts & 12 \\
			4 & ButtermilkBiscuits & 54 \\
			5 & Calzone & 41 \\
			6 & CauliflowerFlorets & 13 \\
			7 & CauliflowerSteaks & 8 \\
			8 & Cheesecake & 94 \\
			9 & ChickenDrumsticks & 44 \\
			10 & ChickenNuggets & 36 \\
			11 & ChickenTenders & 14 \\
			12 & ChocoChipCookie & 97 \\
			13 & CinnamonRolls & 62 \\
			14 & Croissants & 169 \\
			15 & CroqueMonsieur & 33 \\
			16 & FrozenCrinkleCutChips & 56 \\
			17 & FrozenOvenChips & 124 \\
			18 & FrozenPizza & 177 \\
			19 & FrozenPizzaSlices & 15 \\
			20 & FruitPie & 35 \\
			21 & Madeleines & 18 \\
			22 & Meatballs & 16 \\
			23 & Muffin & 50 \\
			24 & NuggetsWithOvenChips & 36 \\
			25 & PizzaClassicCrust & 162 \\
			26 & PotatoChunks & 50 \\
			27 & PotatoWedges & 144 \\
			28 & SalmonSteak & 57 \\
			29 & TaterTots & 26 \\
			30 & Tortillas & 16 \\
			\bottomrule
		\end{tabular}
	\end{adjustbox}
	\caption{Recipe dataset with total number of cooking sessions.}
	\label{tab:recipes}
\end{table}

\begin{figure*}[ht]
	\centering
	\includegraphics[width=0.9\linewidth]{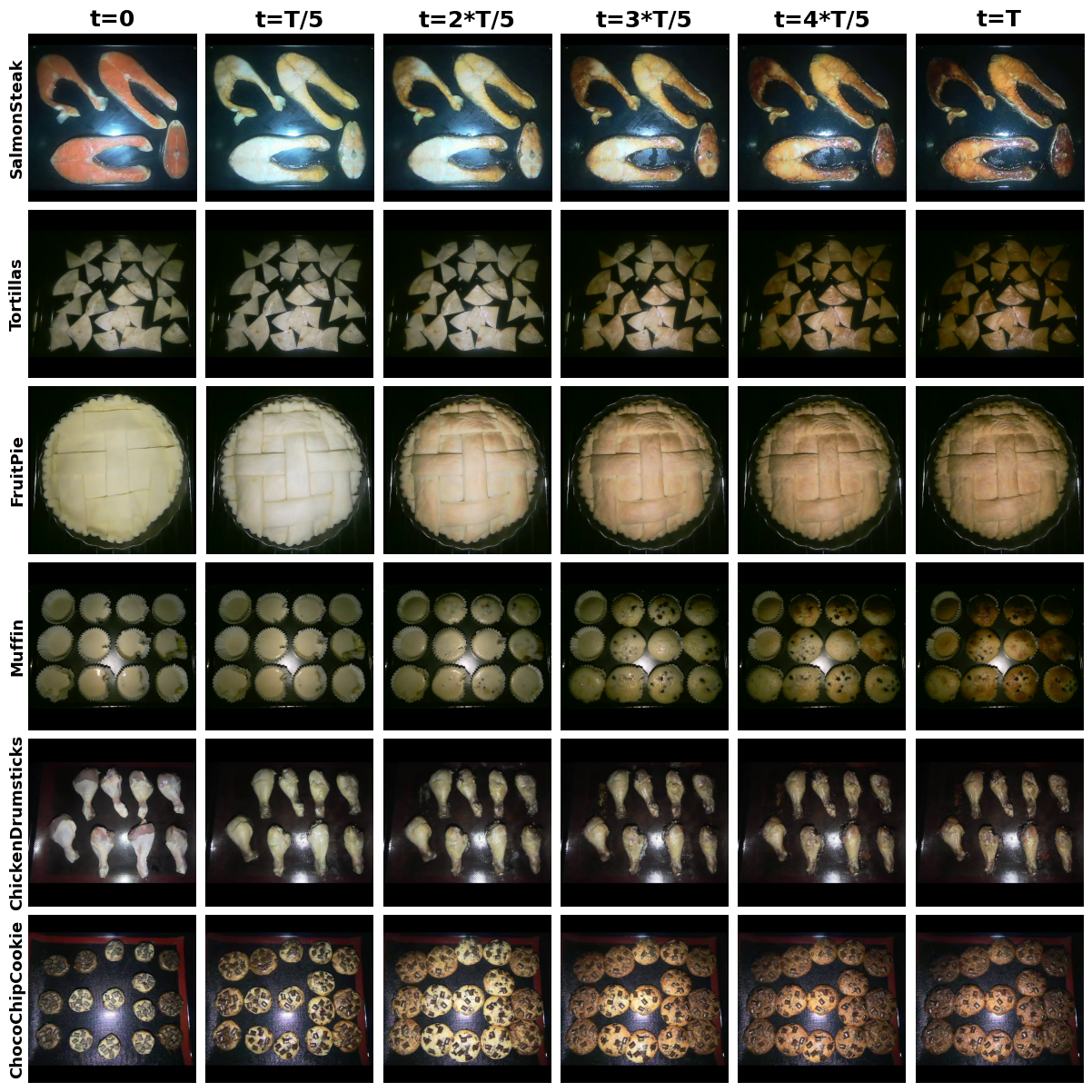}
	\caption{Few samples from cooking session data; t=0 shows raw image and t=T shows last image. In between we show equally sampled images from the session. }
	\label{fig:session}
\end{figure*}

\section{Experimental setup for Cooked Food Image Generation}
\label{appendix:experimentgen}
\subsection*{Generator Architecture}
The overall system processes $224 \times 224$ RGB images with 3 input/output channels and supports text guided conditional image generation.
The generator uses a custom U-Net architecture with a base dimension of 32 and dimension multipliers of $(1, 2, 4, 8)$ for multi-scale feature extraction. It consists of 4 downsampling blocks, 4 upsampling blocks, and 2 middle processing blocks. Each downsampling block contains two ResNet blocks with 8 groups each, followed by improved downsampling, while upsampling blocks include two ResNet blocks with skip connections and upsampling operations. The generator incorporates recipe name and cooking state embeddings with a $4\times$ expanded dimension (128) processed through a SiLU activation network. Text guidance is achieved through 32-dimensional sinusoidal positional embeddings with $\theta = 10000$, supporting dynamic mapping of 30+ food categories. The network uses residual connections throughout and concludes with a final residual block and $1 \times 1$ convolution for output generation.

\subsection*{Discriminator Architecture}

The discriminator employs a $70 \times 70$ PatchGAN classifier with batch normalization, operating on concatenated input and output images (6 channels total). It uses a progressive architecture starting with 64 filters, increasing by powers of 2 up to 8 layers, with $4 \times 4$ kernels, stride 2, and padding 1. The network consists of convolutional layers with LeakyReLU (0.2) activation, producing a single-channel prediction map that classifies overlapping image patches as real or fake. This patch-level design enables fewer parameters and fully convolutional operation on arbitrarily-sized images. 

\subsection*{Training Details}

The model was trained for 100 epochs total (50 epochs at initial learning rate + 50 epochs linear decay) using the Adam optimizer with learning rate $2 \times 10^{-4}$, $\beta_1 = 0.5$, and batch size 1. The loss function combines multiple components: vanilla GAN loss ($\lambda_{\text{GAN}} = 1$), LPIPS perceptual loss using VGG network ($\lambda_{\text{LPIPS}} = 50$), and custom $CIS$ loss ($\lambda_{\text{CIS}} = 50$). 

\section{Experimental setup for Culinary Image Similarity Metric}
\label{appendix:experimentCIS}
\subsection*{Network Architecture}
The system processes $224 \times 224$ RGB images. The system utilizes EfficientNet-B1 as the primary backbone architecture. The final classifier layer is replaced with a custom 2048-dimensional linear embedding projection that includes L2 normalization for stable training. For our experiments, we have taken trained EfficientNet-B1 from scratch.
The projection module consists of a two-layer MLP (2048 $\rightarrow$ 2048 $\rightarrow$ 128 dimensions) with ReLU activation, followed by final L2 normalization.

\subsection*{Training Details}
The model was trained for 100 epochs using Adam optimizer with learning rate $1 \times 10^{-4}$ and weight decay $1 \times 10^{-5}$, employing  step decay ($\gamma = 0.6$ every 10 epochs). The training uses a custom MSELoss that computes cosine similarity matrices between normalized embeddings and compares them against ground truth temporal matrices representing cooking progression. Data augmentation includes horizontal flips ($p = 0.5$) and rotation ($\pm 60^{\circ}$). The system uses a batch size of 32 with session-based sampling, where each batch contains temporally ordered images from the same cooking session, enabling the model to learn temporal relationships in food appearance during cooking.

\section{On-device Deployment}
The entire pipeline is designed for real-time, edge-device operation.
The trained models are ported to a ARM-based embedded processor with an inbuilt camera feed.
\begin{itemize}[noitemsep]
	\item The generator $G_\theta$ and similarity network $f_{sim}$ is implemented with PyTorch and exported to ONNX for edge deployment..
	\item The full model stack (generator $+$ similarity network) occupies 45 MB after hybrid quantization(float16+int8).
	\item Image generation runs at 1.2s/frame on a 5 TOPS NPU. So to generate 3 images from a raw image, 3.6s secs are taken. It takes 330ms to compute similarity for each pair (real+generated) on NPU.
\end{itemize}
The system requires no internet connection and integrates into smart appliances with minimal hardware changes, enabling visually aware cooking automation for everyday users.

\section{Additional Results}

\subsection{Gemini generated images}
\label{appendix:gemini}
\begin{figure}
	\centering
	\includegraphics[width=\linewidth]{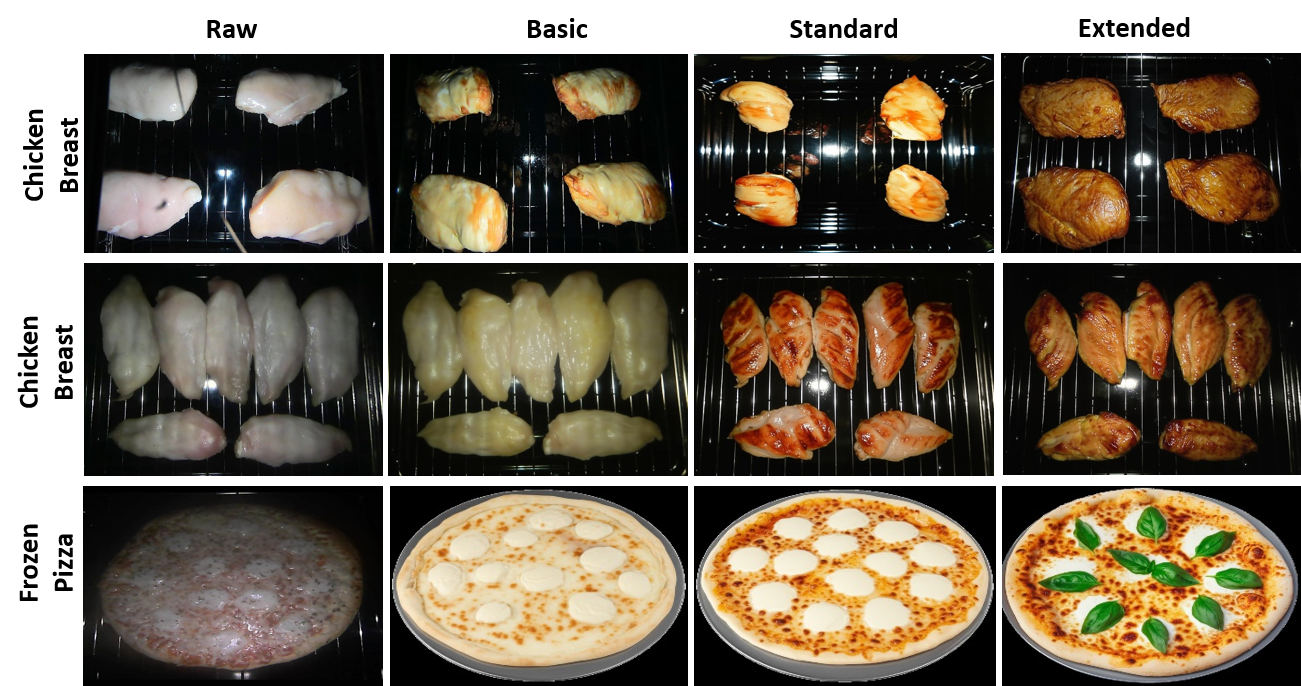}
	\caption{Images generated by Gemini with few shot prompting.}
	\label{fig:gemini}
\end{figure}

We tested the Gemini free API (Gemini-2.0-flash-exp-image-generation) for few-shot generation of cooked-state images. Although visually appealing (\ref{fig:gemini}), the outputs lacked temporal coherence and failed to capture cooking progression reliably (e.g., shifting positions in chicken breast samples, extra basil in pizza images). These inconsistencies limit applicability for real-time cooking state monitoring. Additionally, reliance on cloud deployment introduces potential privacy concerns. 

\subsection{Cooking Progress Monitoring}
\label{appendix:cpm}
Once the user selects a desired cooking state from the three generated options, the cooking process initiates. Frames are captured at 30-second intervals throughout the cooking duration. Each captured frame is compared with the user-selected generated image, and the similarity metric $\mathcal{F}_{cul}$ is computed and stored in a similarity list. After applying smoothing to this list, a sliding window is used to identify local peaks within the window. Upon detecting a peak, the cooking process is automatically stopped. This mechanism enables users to precisely control the cooking process to their desired level, eliminating the need for manual adjustments of oven time and temperature.
\begin{figure}
	\centering
	\includegraphics[width=\linewidth]{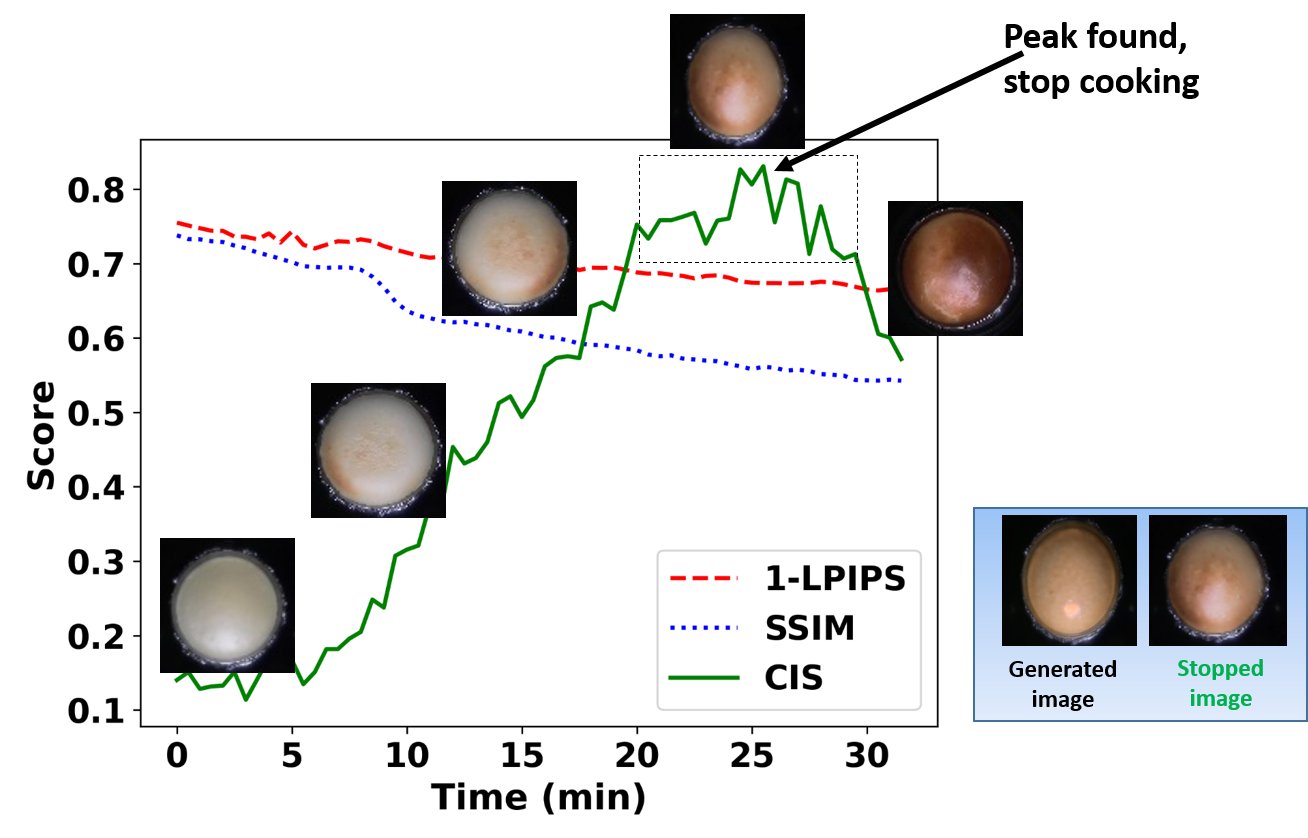}
	\caption{User selects a desired cooking state and cooking begins. Generated and real images are compared at each frame capture. When peak is found, cooking stops.``Stopped image'' is the real image which is closest to the generated image.}
	\label{fig:cookingprogression}
\end{figure}

\subsection{Classification model for doneness detection}
\label{appendix:doneness}
To establish baseline performance for doneness detection, we experimented with three model architectures:
\begin{itemize}[noitemsep]
\item Frame-wise 2D-CNN  processing each image independently.
\item Stacked 2D-CNN aggregating features across frames.
\item 3D-CNN incorporating temporal information explicitly.
\end{itemize}

We evaluated these models on 5 distinct recipes, each annotated with 3 doneness states (e.g., rare, medium, well-done). Despite these efforts, the models achieved only $\approx$50\% accuracy, highlighting their limited generalization capability. This performance gap stems from inherent challenges such as variations in food thickness, moisture content, seasoning distribution, and oven-specific dynamics. These results underscore the need for more robust, instance-aware approaches, such as our proposed generative framework to handle real-world cooking variability effectively.

\end{document}